\def\year{2020}\relax
\documentclass[letterpaper]{article} 
\usepackage{aaai20}  
\usepackage{times}  
\usepackage{helvet} 
\usepackage{courier}  
\usepackage[hyphens]{url}  
\usepackage{graphicx} 
\usepackage{graphicx}  
\usepackage{xcolor, colortbl}
\usepackage{subfig}

\title{How Robust Are Graph Neural Networks to Structural Noise?}
\author{\\ \Large \textbf{James Fox,\textsuperscript{\rm 1,2} Sivasankaran Rajamanickam\textsuperscript{\rm 2} }\\ 
\textsuperscript{\rm 1}Georgia Institute of Technology\\ 
\textsuperscript{\rm 2}Center for Computing Research, Sandia National Laboratories\\
jfox43@gatech.edu, srajama@sandia.gov
}

\begin{document}

\maketitle

\begin{abstract}
Graph neural networks (GNNs) are an emerging model for learning graph embeddings and making predictions on graph structured data. However, robustness of graph neural networks is not yet well-understood. In this work, we focus on node structural identity predictions, where a representative GNN model is able to achieve near-perfect accuracy. We also show that the same GNN model is not robust to addition of structural noise, through a controlled dataset and set of experiments. Finally, we show that under the right conditions, graph-augmented training is capable of significantly improving robustness to structural noise. 
\end{abstract}

\section{Introduction}
\label{sec:intro}
Using graph embeddings for predictions on graph-structured data has been growing in recent years. One particular class of models is the Graph Neural Networks (GNN), with early versions inspired by extension of convolutional neural networks to graph based approaches. Such models have led to state-of-the-art classification on benchmark datasets, and are being actively applied to other diverse domains. 

However, robustness of GNN models has not been well-studied, nor strategies for improving their robustness, compared to strategies for robustness in other domains such as natural language processing and computer vision. GNN robustness has been studied somewhat in the adversarial setting, edge insertions and/or removals are algorithmically determined.  Our work focuses on structural noise in a non-adversarial, random model. 

Our approach is to start with understanding robustness of GNNs in the structural prediction setting, where GNNs have been shown to be theoretically quite powerful \cite{xu2018powerful}--as powerful  as the Weisfeiler-Lehman graph isomorphism test \cite{shervashidze2011weisfeiler}. Empirically, we show that the GNN proposed in \cite{xu2018powerful}, the Graph Isomorphism Network, is able to near-perfectly classify nodes defined strictly based on structural identity. We use this particular GNN as it was shown to be theoretically maximally powerful among several variations. We refer to it as GIN in the remainder of the paper.

We evaluate GIN robustness in terms of its ability to train and predict ground truth labels in the presence of noise. We define noise as randomly added edges that change graph structure but not ground truth labels. We find that the accuracy of the model rapidly declines from adding up to 25\% noise with respect to number of edges of the original graph, dipping below 50\% F1-score at 25\% noise. Improving the number of training samples from the noisy graph recovers some performance, but with severely diminishing returns.

To this end, we also explore augmenting training with generated noisy graphs in a manner same to drawing a noisy graph from the same distribution to add to training. We also consider augmenting with noisy samples generated from smaller versions of the original graph, which do not necessarily have same noise distribution as in the original graph. We show that augmenting training with generated noisy samples can significantly improve robustness, whether with noisy samples with the same ground truth labels or with smaller augmented graph samples. 

To summarize contributions, we (1) empirically show that GINs are capable of perfectly distinguishing structural identity, with minimal training samples (2) GIN accuracy sharply declines with structural noise as random edge additions, even when those additions are distance-restricted (3) we show that under certain conditions, augmented training with generated samples can be a powerful tool for improving GNN model accuracy, even when the target data is structurally noisy.


\section{Related Work}
\subsection{Node Structural Identity and Prediction}
\cite{henderson2012rolx}, \cite{ribeiro2017struc2vec}, and \cite{donnat2018learning} focus on learning unsupervised embeddings for distinguishing and classifying nodes by their structural role or identity in a graph. \cite{xu2018powerful} also connects the capacity of a class of GNNs to their ability to distinguish structures in terms of multi-hop neighborhoods.

\subsection{Graph Neural Networks}
Several works introduced the notion of graph convolutions into models for learning embeddings of graph-structured data, such as \cite{kipf2016semi} and \cite{defferrard2016convolutional}. Many other GNN models and variations have since been proposed, such as \cite{gilmer2017neural} with a message passing perspective, \cite{velivckovic2017graph} with attention, \cite{hamilton2017inductive} with sampling. \cite{xu2018powerful} provided a theoretical framework for generalizing several GNNs, and also propose an optimal model under this framework. 

\subsection{Robustness of Graph Neural Networks}
\cite{dai2018adversarial} propose a reinforcement learning approach, as well as several non-learning based algorithms, to make edge modifications to graph structure. In all of their adversarial attack settings, the model is considered fixed and is not re-trained in the presence of new structural modifications. 

\cite{zugner2018adversarial} consider adversarial modifications of both graph structure and features, by adopting a greedy approach and a surrogate model scheme designed to be computational tractable. While our setting is not adversarial, similar to our work they evaluate when when a model is retrained following modifications ( referred to as ``poisoning attack'' in their work).

\section{Methods}
In this section we focus on the overall methodology for evaluating robustness of the GIN model to structural noise. We define the noise model, as well as the synthetic graph used for evaluation. We also present an augmented training strategy to improve robustness, under conditions that new labeled samples can be generated from the same noisy distribution, serving as an upper bound for relevance of generated samples in dataset. Furthermore, we extend the augmented training to noisy samples generated from smaller versions of the original graph. We introduce symbols and notations as needed in subsequent subsections.  

\subsection{Graph Generation}
\begin{figure}[t]
	\center 
	\includegraphics[width=50mm]{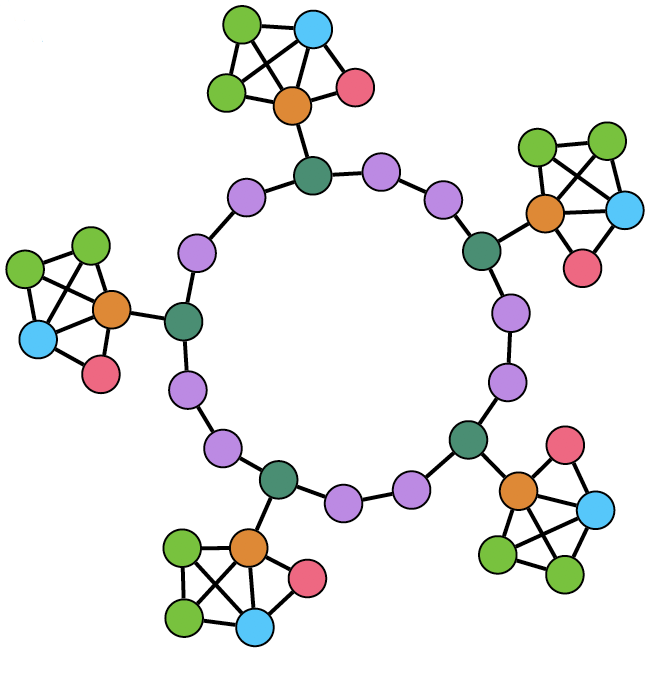}
	\caption{Graph constructed as ring of houses. Concept and image from \cite{donnat2018learning}.}
	\label{fig:ring-of-houses}
\end{figure}

We used the graph generation concept from \cite{donnat2018learning} to create a graph with ground-truth structural identity labels. 
In our work, we chose a ring of nodes with identical ``house" motifs attached (see Fig. \ref{fig:ring-of-houses}). We refer to this graph as $G$. While other choices of graphs are possible, in this graph the nodes' visual structural identities are also verifiable through the procedure of the Weisfeiler-Lehman isomorphism test.

As seen in Fig. \ref{fig:ring-of-houses}, $G$ has 6 classes corresponding to 6 different structural roles. It is trivial to upsize or downsize this graph while maintaining the same number of classes. We use a ring of 333 houses attached at regular intervals to a ring of 999 nodes (unless noted otherwise), as well as a smaller version with 264 nodes and 396 edges in later experiments. More details on the graphs are found in Table \ref{tab:dataset}. 

\begin{table}[t]
	\caption{Ring of houses dataset}
	\begin{tabular}{|l|l|l|l|l|}
		\hline
		Name & Nodes & Edges & Classes \\ \hline
		Ring of houses $G$ & 2664 & 3996 & 6 \\ \hline
		Ring of houses $G'$ & 264 & 396 & 6 \\ \hline
	\end{tabular}
	\label{tab:dataset}
\end{table}

\subsection{Noisy Graph Generation}
\label{sec:noise-gen}
In general, node labels do not perfectly match the structure of their neighborhood--i.e., nodes with different neighborhoods can belong to the same class. We aim to randomly add noise in the form of newly sampled edges to $G$ to introduce a mismatch between structure and labels in a controlled manner. The labels remain exactly the same; the only difference is in graph structure. 

Let $G_p$ denote the noisy version of $G$ with added structural noise ratio $p$. We model random noise as edge additions in two different settings: global, and distance-limited. In the global setting, new edges are created from uniformly sampling node pairs at random (without replacement). The distance-limited setting allows us to define a notion of more localized noise and restrict edge additions accordingly. In the distance-limited setting, we limit new edges to form only between pairs that are within $k$-hop distance of each other.
When new edges are added, we update features to match new degrees.

\begin{figure*}%
	\centering
	\subfloat[No noise]{{\includegraphics[width=57mm]{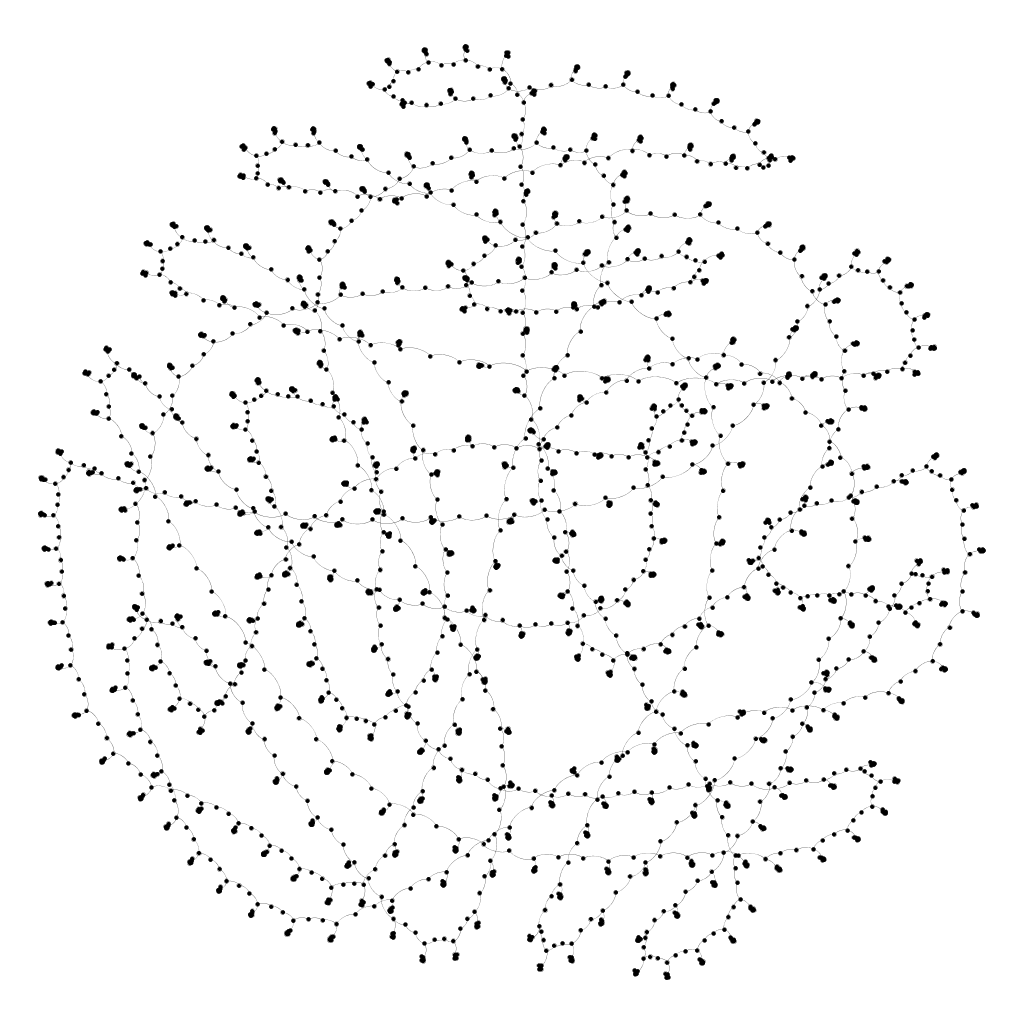} }}%
	\subfloat[Within 2-hop noise]{{\includegraphics[width=57mm]{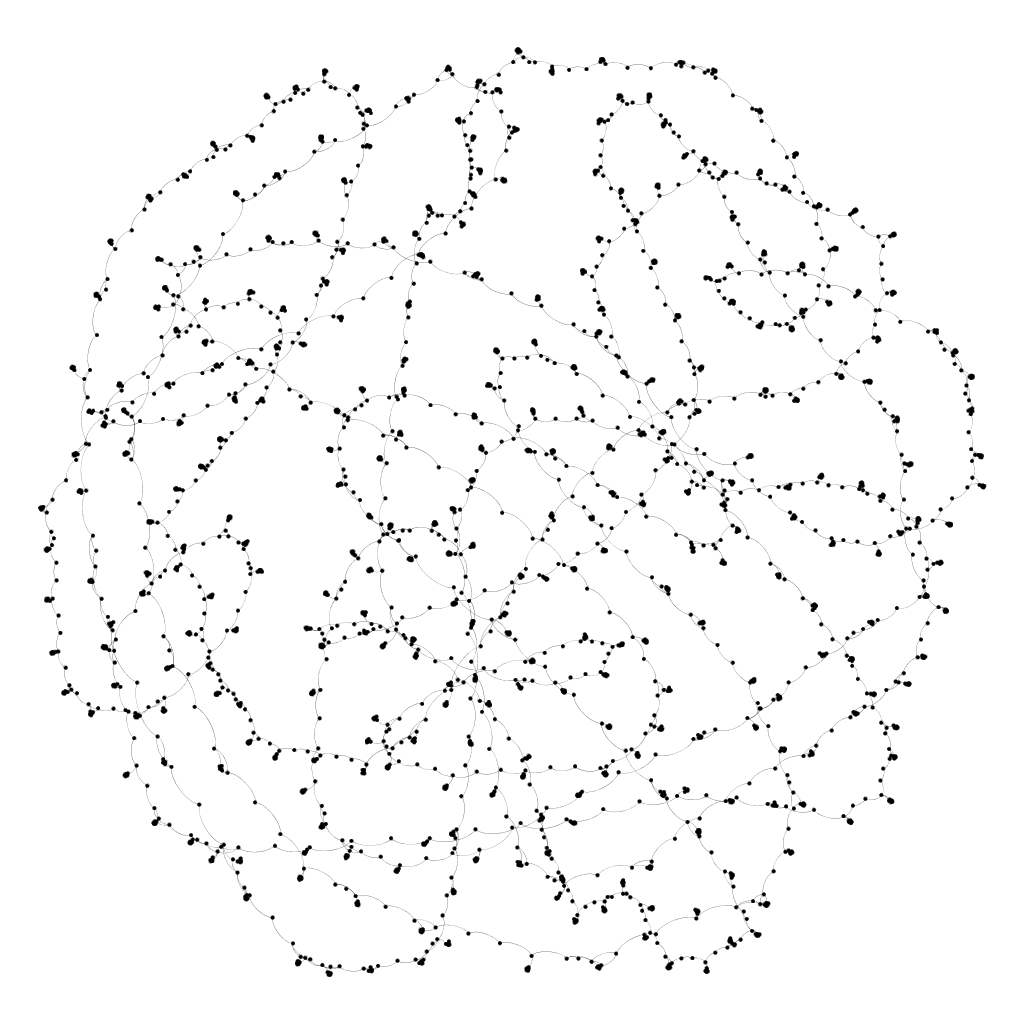} }}%
	\subfloat[Within 3-hop noise]{{\includegraphics[width=57mm]{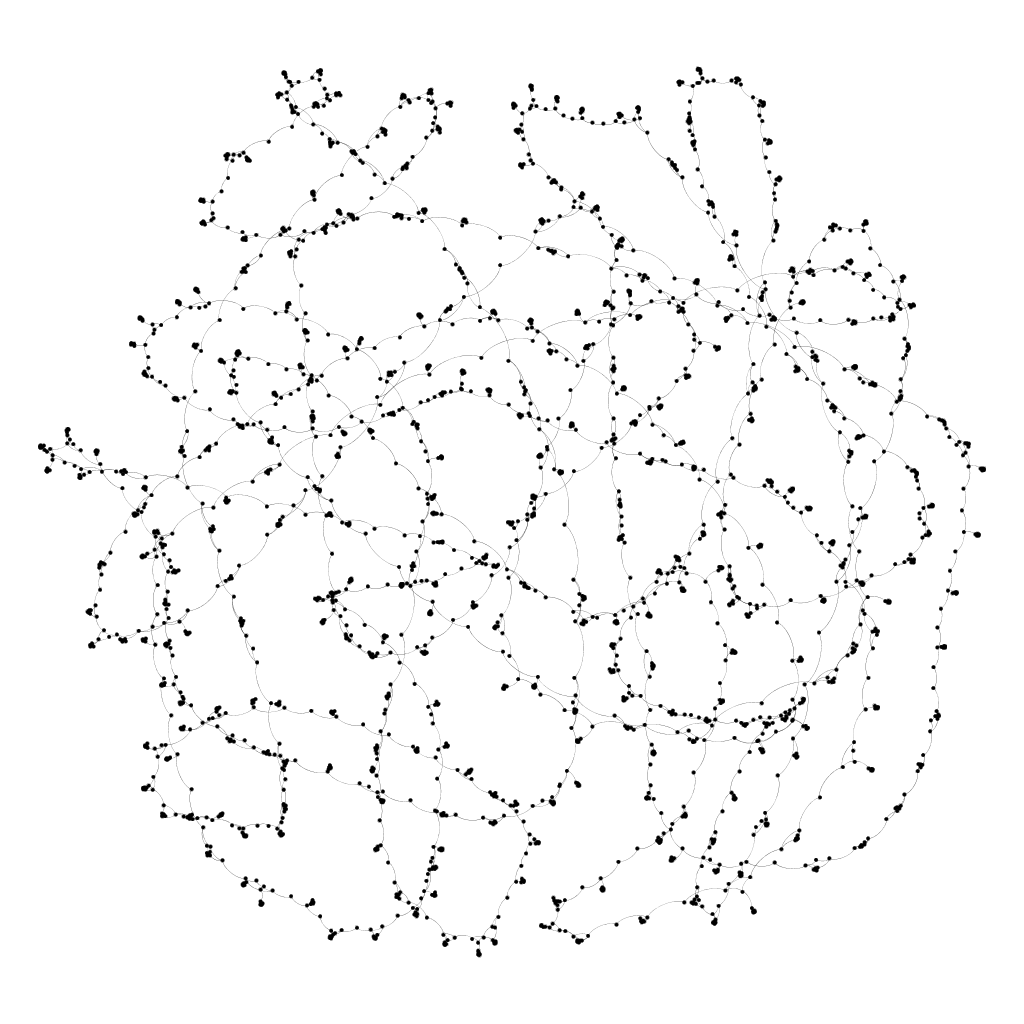} }}%
	\caption{Gephi visualization of $G$, and 2-hop and 3-hop limited noisy $G_p$ versions using $p=0.15$ noise ratio.}%
	\label{fig:gephi}%
\end{figure*}

We refer to Fig. \ref{fig:gephi} for visualizations of full 2-hop and 3-hop limited graphs (noise ratio fixed at $p=0.15$) generated using Gephi \cite{bastian2009gephi} visualization tool with Force Atlas layout and default parameters. We chose the layout that gave good trade-off between intuitive visualization and generation time.

While we only consider edge additions, robustness to edge removals would be an interesting avenue of future work. It would be necessary to consider new noise models in this setting, as a local (distance-limited) analogue for edge-removal has to be developed. Early studies of  complex networks such as \cite{albert2000error} show simple edge removal models can lead to high frequency of tiny clusters (singletons or size two clusters), which would be challenging for GNN structural prediction.

\subsection{Augmented Training for Improving Robustness}
In many cases, the number of training samples available in a graph is a limited resource. Obtaining more training samples from an existing graph is not always feasible. This motivates exploring the impact of augmenting a structurally noisy graph $G_p$ with generated samples to improve robustness. 
We explore sample graph generation in two cases: when the sample is drawn from the same distribution as $G_p$, and when it it's not drawn from the same distribution but still very related. 

The former represents an ideal data augmentation scenario. Let $G_p^{(j)}$ refer to the $j$-th graph generated from same noise distribution as $G_p$, used in augmented training. Specifically, in our setting we generate graph $G_p^{(1)}$ from $G$ using the same procedure used to generate $G_p$ from $G$. In other words, $G_p^{(1)}$ is drawn from the same distribution as $G_p$.

We also consider the case where $G_p^{(1)}$ is not drawn from the exact same distibution as $G_p$, by generating a smaller graph version $G_p'^{(1)}$. $G_p'^{(1)}$ is generated from $G'$, which is $10\times$ smaller than $G$ in terms of number of vertices and edges (see Table \ref{tab:dataset}. Therefore the noise ratio is also with respect to the smaller graph. $G'$ still maintains the same structural identities as in $G$, and so the distribution of labels remains unchanged.

\section{Experiments and Results}

\subsection{Model}
We use the Graph Isomorphism Network from \cite{xu2018powerful} as the GNN. We implement the GIN using PyTorch Geometric \cite{fey2019fast} library. Our model uses 3 GIN layers, followed by two fully connected layers (the last of these two is output). Hidden dimension is 32 across all weights. 
Each GIN layer contains two fully connected layers (applied after aggregation). Every GIN layer is accompanied by a Batch normalization layer, which proves quite beneficial to classification performance in our case. ReLU activation is used after linear  transformations and GIN aggregation. The only feature used is node degree, which is normalized with zero mean and unit standard deviation. 
For optimization, we use Adam with learning rate of $0.01$ and weight decay of $5\times10^{-4}$.


\subsection{Experiment Setup}
In all experiments, we record results from 50 independent trials. For each trial, unless specified otherwise, we use 20 samples per class from $G_p$ as training set (total of 160 nodes), 200 total nodes for validation, and 1000 nodes for testing. 
Training, validation, and test splits are fixed according to random seed. 
We use predetermined seeds for all the synthetic graphs used in our experiments for data reproducibility.
Model optimization and methods internal to PyTorch may be a source of randomness. 

We evaluate predictive performance in terms $F_1$-macro score. While there is some class imbalance, it is not significant, and we treat classes as equally important. For simplicity, we refer to this as just the $F_1$ score. 
We take the test score corresponding to the model achieving best validation score while training for 200 epochs. 

\subsection{Baseline Noise}
\begin{figure}[h!]
\center 
\includegraphics[width=80mm]{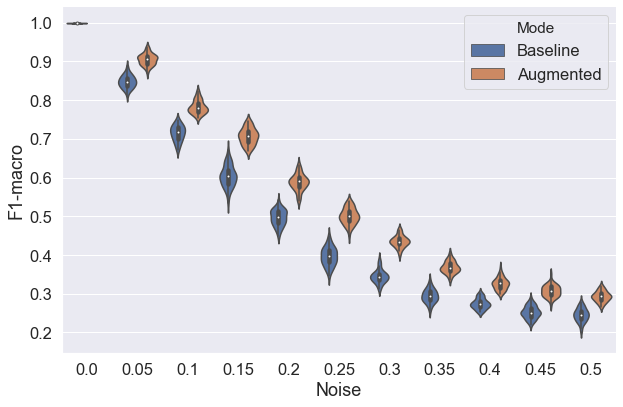}
\caption{Test F1 score of GIN model with varying levels of structural noise added to input graph, across 3 different modes of noise constraint. Model is re-trained after noise addition.}
\label{fig:baseline-noise}
\end{figure}

We vary the ratio $p$ of noisy edges added to $G$ (with respect to number of edges in $G$) in increments of 0.05, and examine the performance of the model when trained on $G_p$ for respective $p$. Edges are added in the manner described in earlier section, over 3 different modes of noise: constrainted to 2-hop, 3-hop, or unconstrained (global). 20 nodes per class are used for training in all trials. 

Results of testing on 1000 nodes while varying the noise from 0.05 to 0.5 are shown in Fig. \ref{fig:baseline-noise}. Each violin plot gives the distribution from 50 trials. When there is no noise added, i.e. $G = G_p$, the GIN model learns to classify nodes near perfectly. This suggests that the GIN can be empirically very good at distinguishing structural identity. In fact, we found just one node sample per class was sufficient to train $G$ to convergence in test accuracy. 

However, the F1-score across all modes declines sharply with the introduction of randomly added edges. This means that the GIN is not able to recover labels with added structural noise nearly as well. At $p=0.25$, the average performance across all modes drops below 50\%.

There are some additional trends of interest. Across all modes, performance begins to flatten out from $p=0.3$ onwards. This could indicate some limit beyond which there is too much noise such that it becomes very difficult to distinguish nodes by structure alone. For instance, if nodes in the house form cliques, then they would be indistinguishable up to some model depth. Also, the performance curves for 2-hop, 3-hop, and global noise seems to diverge around $p=0.25$ onwards, at which point the 3-hop mode shows worst performance. These trends remain to be investigated further. 

\subsection{Varying Number of Training Samples}
\begin{figure}[h!]
\center 
\includegraphics[width=80mm]{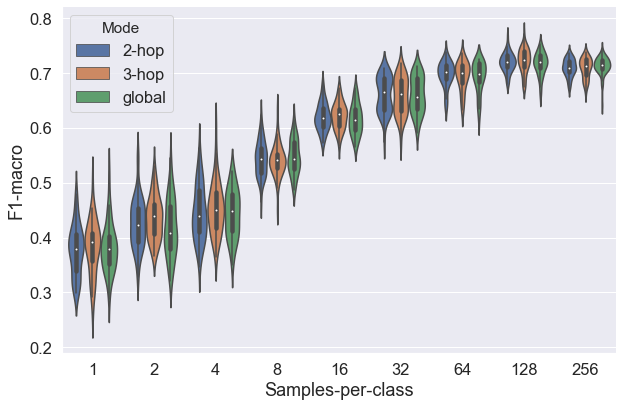}
\caption{Test F1 score vs. number of node labels per class available for training on $G_p$. Noise is fixed at $p=0.15$.}
\label{fig:samples}
\end{figure}

\begin{figure*}%
	\centering
	\subfloat[$2$-hop noise]{{\includegraphics[width=57mm]{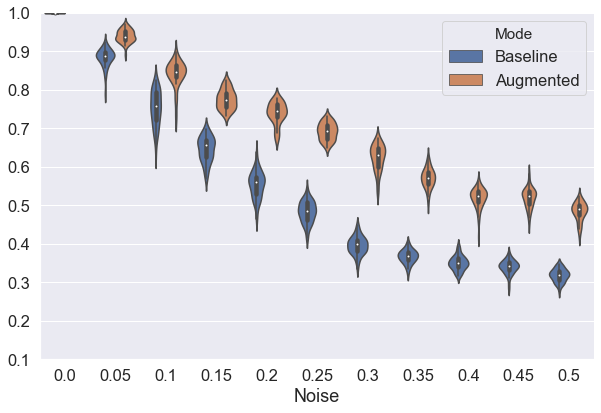} }}%
	\subfloat[$3$-hop noise]{{\includegraphics[width=57mm]{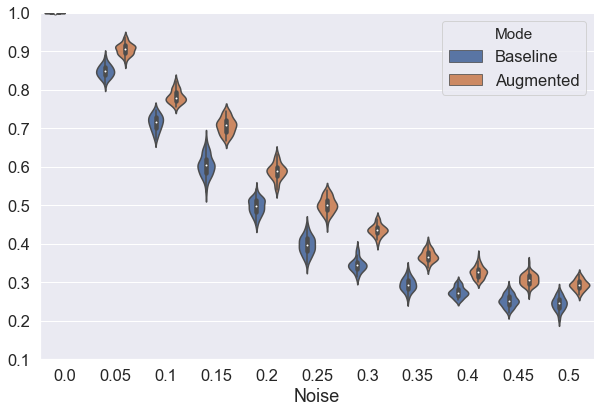} }}%
	\subfloat[Global noise]{{\includegraphics[width=57mm]{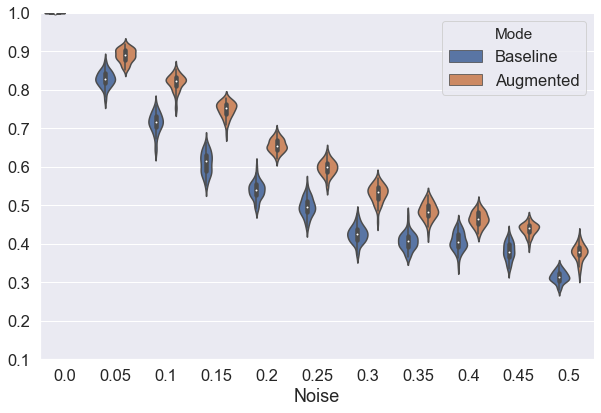} }}%
	\caption{Augmented vs. non-augmented training (baseline) for node classification on $G_p$. Y-axis is F1 score, x-axis is random edge addition ratio.}%
	\label{fig:augment}%
\end{figure*}

\begin{figure*}%
	\centering
	\subfloat[$2$-hop noise]{{\includegraphics[width=57mm]{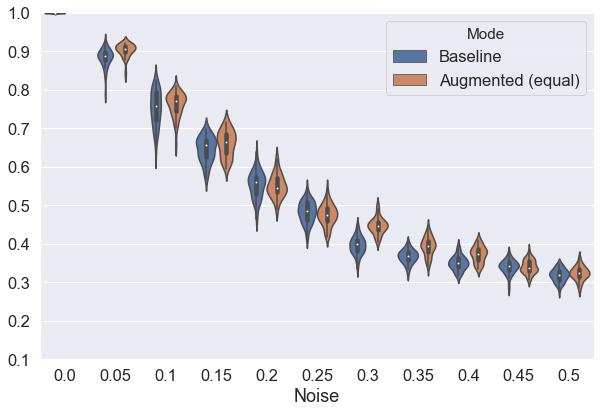} }}%
	\subfloat[$3$-hop noise]{{\includegraphics[width=57mm]{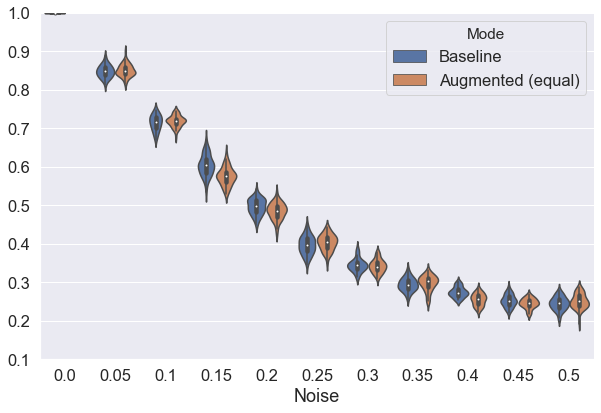} }}%
	\subfloat[Global noise]{{\includegraphics[width=57mm]{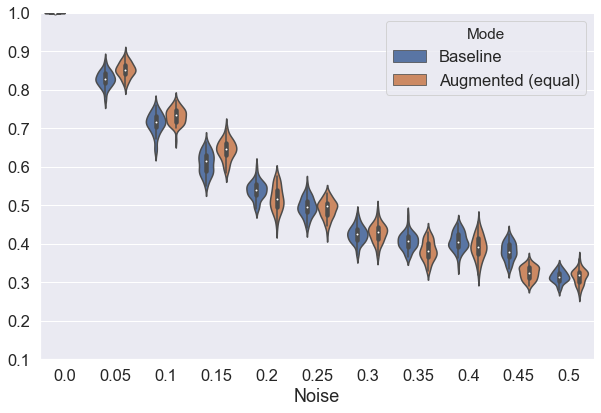} }}%
	\caption{Augmented vs. non-augmented training (baseline) with equal number of total training samples in each setting (20 samples-per-class). Node classification is on $G_p$.}%
	\label{fig:augment-eq}%
\end{figure*}

\begin{figure*}%
	\centering
	\subfloat[$2$-hop noise]{{\includegraphics[width=57mm]{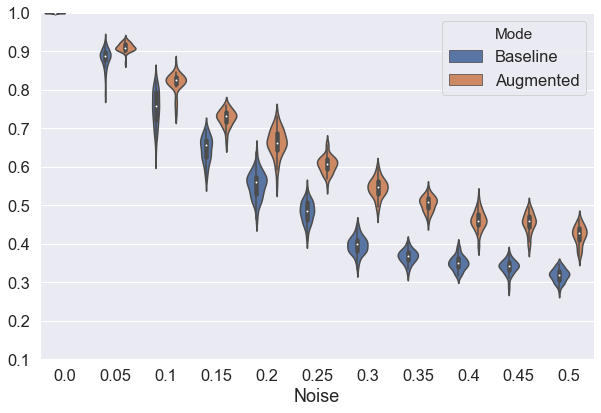} }}%
	\subfloat[$3$-hop noise]{{\includegraphics[width=57mm]{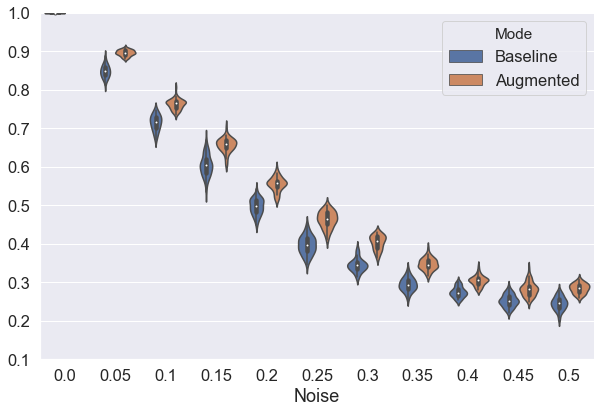} }}%
	\subfloat[Global noise]{{\includegraphics[width=57mm]{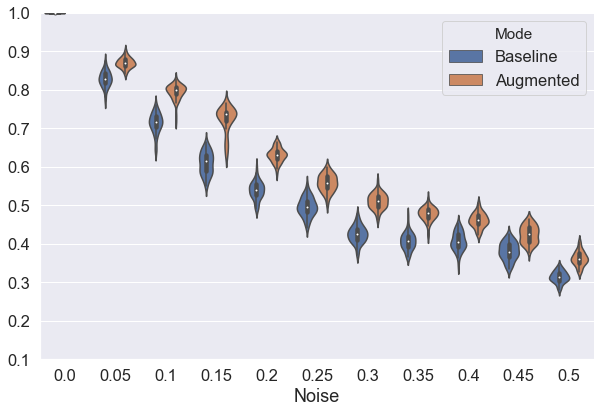} }}%
	\caption{Augmented training with multiple smaller graphs vs. non-augmented training (baseline) for node classification on $G_p$. Y-axis is F1 score, x-axis is random edge addition ratio.}%
	\label{fig:augment-small}%
\end{figure*}

Instead of fixing the number of node training labels per class at 20, we vary this parameter in powers of 2, from 1 up to 256 training labels per class. 
We fix the noise level at $p=0.15$ for 2-hop, 3-hop, and global noise modes, and show results in Fig. \ref{fig:samples}. Increasing the number of training samples in $G_p$ improves performance up to around 64 samples-per-class, after which point there is severely diminishing returns even with exponentially growing number of samples. This trend is essentially mirrored across all modes of the noisy graph.

The results suggest that there is a limit to which the model can recover labels from a structurally noisy graph, that having more training samples from the same graph won't necessarily help. 
For $G_p$ with $p=0.15$, this limit is between 70-72\% F1 score (median), reached between 64 and 128 samples per class. This also corresponds to training labels for 14\% and 28\% of total number of nodes of $G_p$, which is quite a high demand.

\begin{table}[h!]
\begin{center}

\caption{Percentage median F1 score improvement over baseline from (a) augmented training with graph from same noise distribution, and (b) smaller graph versions.}
\small

\subfloat[Augmentation with $G_p^{(1)}$]{\begin{tabular}{|c|c|c|c|} \hline
Noise Ratio & 2-hop & 3-hop & Global \\ \hline
0.05 & 5.63& 6.98& 7.35\\   \hline
0.1 & 11.7& 8.57& 14.99 \\   \hline
0.15 &18.11 & 16.88& 22.72 \\   \hline
0.20 & 32.79& 18.48& 21.15\\   \hline
0.25 & 42.73& 26.09& 20.67\\   \hline
0.30 & 58.67& 26.39& 25.82\\   \hline
0.35 & 54.85& 24.69& 18.55\\   \hline
0.40 & 50.16& 19.77& 15.1\\   \hline
0.45 & 52.8& 22.0& 16.3\\   \hline
0.5 & 53.23& 19.35 & 21.06 \\   \hline
\end{tabular}}
\quad
\subfloat[Augmentation with $G_p'^{(j)}$]{\begin{tabular}{|c|c|c|c|} \hline
Noise Ratio & 2-hop & 3-hop & Global \\ \hline
0.05 & 2.48 & 5.64& 4.81 \\   \hline
0.1 & 8.78& 6.78& 11.66 \\   \hline
0.15 & 11.42& 8.9& 19.93  \\   \hline
0.20 & 18.19&12.04 & 16.99\\   \hline
0.25 & 25.2& 17.19& 12.25\\   \hline
0.30 & 37.57& 18.5 & 20.33 \\   \hline
0.35 & 37.72& 17.57& 17.85 \\   \hline
0.40 & 31.85& 11.9& 14.08 \\   \hline
0.45 & 33.37& 12.5& 11.89\\   \hline
0.5 & 33.94& 15.8 & 14.93\\   \hline
\end{tabular}}
\label{tab:augment}
\end{center}
\end{table}

\subsection{Augmented Training}
Augmented training aims to add new data to training from outside the existing dataset. 
We first consider a graph generated from the same distribution as $G_p$. The augmented training procedure proceeds by training on both $G_p^{(1)}$ and the $C*l$ samples of $G_p$, where $l$ is number of samples-per-class (fixed at 20) and $C$ is the number of classes ($C=6$ in all cases). All node labels of $G_p^{(1)}$ are used for training. In every epoch, we first train on $G_p^{(1)}$ and then train on the samples of $G_p$. Validation and testing is performed on $G_p$ only, hence $G_p^{(1)}$ only augments the training data.

Results from this augmented training are presented in Fig. \ref{fig:augment}. Compared to using just 20 samples-per-class for regular training, the augmented version significantly improves F1 score in all cases. $G_p^{(1)}$, drawn from the same distribution as $G_p$, contributes an additional 2664 nodes for training to augment the original 120 total samples of $G_p$. 

Training augmentation with a graph drawn from the same noise distribution results in relative improvement of median F1 score up to 59\%, 26\%, and 26\% for 2-hop, 3-hop, and global noise modes, respectively. We refer to Table \ref{tab:augment} for full results. 

While the total number of training samples in the augmented case (drawn from $\{G_p, G_p^{(1)}\})$ is signficantly higher than that of the baseline (drawn from $G_p$) by design, we also directly compare training on $G_p$ vs. $G_p^{(1)}$ using identical training setings. Using 20 samples-per-class, we independently train one model using $G_p$ and another using $G_p^{(1)}$, and test both on $G_p$. Results, shown in Fig. \ref{fig:augment-eq}, verify the quality of the augmenting graph. Training on $G_p^{(1)}$ results in similar model performance as training on $G_p$, even though test nodes are from $G_p$. Across all modes, relative improvement from training on the augmented graph varies from $12\%$ to $-14\%$, which indicates variation in graph instances from same noise distribution.

Next, we consider smaller graphs $[G_p'^{(1)}, G_p'^{(2)},...,G_p'^{(n)}]$, where each $G_p'^{(j)}$ is the noisy graph generated from $G'$, the smaller version of $G$. 
We use $n=10$ in our experiments. We adapt the augmented training procedure earlier to handle multiple graphs by treating each as its own batch. We iterate through each batch/graph in random order, and perform augmented training using each graph as one epoch. The batches are reshuffled every $n$ epochs. Results of augmenting with smaller graphs are shown in Fig. ~\ref{fig:augment-small}.

Interestingly, even though the generated graphs are from a different distribution compared to $G_p$, augmented training using them is still beneficial in all cases. The relative improvements are up to 38\%, 18\%, and 20\% for 2-hop, 3-hop, and global noise modes respectively (see Table \ref{tab:augment}). We see similar trends as in Fig. \ref{fig:augment} repeated across the different noise modes, even if the relative improvement with respect to median F1 score is less compared to augmenting training with a graph from the same distribution.

\section{Conclusion}
In this work, we determined that GNN performance can greatly suffer from the addition of random edges as noise in the node classification based on structural identity. We focus on a particular GNN variation, the GIN, in experiments. Across both local and global random noise variations, the GIN  quickly declines in ability to recover labels in the presence of increasing structural noise--even though it achieves near-perfect performance on our dataset prior to introduction of noise.  

To improve robustness to noise, we demonstrated a training augmentation strategy based on generating noisy samples drawn from both the same distribution, as well as a similar but different distribution. This augmentation is quite effective in increasing ability to recover labels from structural noise in most cases, but could be considered empirically an an upper bound in our setting. 

\section{Future Work}
In this work we use a synthetic graph with well-defined structural identity to evaluate the predictive power of GNNs in relation to noise. A direction of future work is to extend our experiments to complex networks, such as Erdos-Renyi or Barabasi-Albert models. As structural identity is ambiguous in this setting, one possibility is to assign WL labels. However, there could be tradeoff between capturing depth of neighborhood vs. label entropy. Another possiblity is to consider non-learned methods for structural embeddings, and clustering in order to produce structural identity labels while controlling label entropy.

Another direction of future work is to study and characterize the relationship between structure and classes in real-world networks. For these networks, labels are not determined by structure alone, as demonstrated by importance of incorporating domain node features. Furthermore, GNNs (without further modifications) do not perform well on primarily non-structural relationships even without noise, such as nodes labeled by communities, which is noted in \cite{you2019position} and also seen in our initial experiments using LFR benchmark graphs \cite{lancichinetti2008benchmark}.

\section{Acknowledgement}
We thank Umit Catalyurek for helpful discussions and revisions. Sandia National Laboratories is a multimission laboratory managed and operated by National Technology and Engineering Solutions of Sandia LLC, a wholly owned subsidiary of Honeywell International Inc. for the U.S. Department of Energy’s National Nuclear Security Administration under contract DE-NA0003525.


\bibliography{main}
\bibliographystyle{aaai}

\end{document}